\documentclass[conference,letterpaper,10pt]{IEEEtran}
\IEEEoverridecommandlockouts
\usepackage{multirow}
\usepackage{amsmath}
\usepackage{amssymb}
\usepackage{graphicx}
\usepackage{booktabs}
\usepackage{array}
\usepackage[table]{xcolor}
\usepackage{tabularx}
\usepackage{makecell}
\usepackage[protrusion=false]{microtype}

\usepackage{enumitem}
\usepackage{url}
\usepackage[hidelinks]{hyperref}
\graphicspath{{./figures/}}

\definecolor{ink}{HTML}{253858}
\definecolor{sizeblue}{HTML}{1F5FBF}
\definecolor{powerteal}{HTML}{00857D}
\definecolor{eventpurple}{HTML}{7655B5}
\definecolor{warmaccent}{HTML}{C2410C}
\definecolor{tablehead}{HTML}{ECF2FA}
\definecolor{sizeband}{HTML}{DCE9FA}
\definecolor{powerband}{HTML}{DCEFEB}
\definecolor{eventband}{HTML}{EAE3F5}
\definecolor{neutralband}{HTML}{E7EBF0}
\definecolor{rowtint}{HTML}{F6F8FA}
\newcommand{\tabhead}{\rowcolor{tablehead}}
\newcommand{\tabnote}[1]{\par\vspace{2pt}\begin{minipage}{\linewidth}\scriptsize #1\end{minipage}}
\newcommand{\res}[2]{\ensuremath{#1{\scriptscriptstyle\,\pm\,#2}}}
\newcommand{\best}[1]{{\bfseries\boldmath #1}}
\newcolumntype{L}{>{\raggedright\arraybackslash}X}
\newcolumntype{C}{>{\centering\arraybackslash}X}

\setlist[enumerate]{leftmargin=*,nosep,topsep=3pt}

\title{VTT: Reading Vehicle Size and Powertrain Type from Motion Trajectory Across Unseen Car Models}

\newif\ifanonymous
\anonymoustrue
\author{%
  \parbox{\textwidth}{%
    \centering
    Yuhang Wang\textsuperscript{1},
    Yuechen Guo\textsuperscript{1},
    Jingxin Yang\textsuperscript{2},
    Chuheng Wei\textsuperscript{3},
    Suyang Chen\textsuperscript{4},\\
    Abdulaziz Alhuraish\textsuperscript{1},
    Yiyi Wang\textsuperscript{1},
    Mingyang Ban\textsuperscript{1},
    Jingran Sun\textsuperscript{1},
    Hao Zhou\textsuperscript{1,*}
  }%
  \thanks{\textsuperscript{1}University of South Florida.}%
  \thanks{\textsuperscript{2}NVIDIA.}%
  \thanks{\textsuperscript{3}Amazon.}%
  \thanks{\textsuperscript{4}University of Florida.}%
  \thanks{\textsuperscript{*}Corresponding author: Hao Zhou.}%
}
\hypersetup{pdfauthor={},pdfsubject={},pdfkeywords={}}

\IEEEaftertitletext{\begin{minipage}{\textwidth}
\centering
\includegraphics[width=\textwidth]{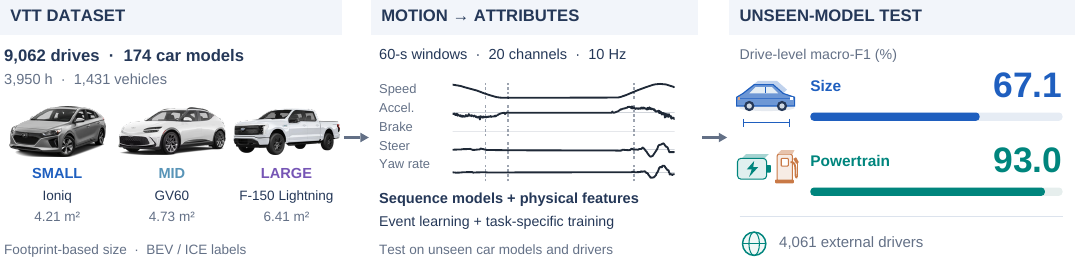}
\makeatletter\def\@captype{figure}\makeatother
\caption{VTT overview. Motion and CAN signals support size and powertrain recognition across unseen car models. The benchmark combines sequence models, physical features and event learning, with external validation and physical analysis. Vehicle images illustrate the size labels; scores are drive-level macro-F1.}
\label{fig:teaser}
\vspace{6pt}
\end{minipage}
}

\begin{document}
\bstctlcite{IEEEexample:BSTcontrol}
\maketitle
\thispagestyle{empty}
\pagestyle{empty}

\begin{abstract}
Vehicle size and powertrain information can enrich trajectory data for traffic analysis and simulation. We present VTT, a dataset and benchmark for identifying these attributes from motion and controller area network (CAN) signals across unseen car models. The dataset contains 9,062 drives, about 3,950 h, from 1,431 vehicles and 174 car models. It provides three footprint-based size classes, powertrain labels, and evaluation on unseen car models and unseen drivers. Using 60-s windows, we compare four sequence architectures, statistical features and physical features proposed by large language models. Training with auxiliary physical targets, data augmentation and task-specific weighting raises drive-level macro-F1 to 92.6--93.0\% for battery-electric versus combustion vehicles and from 62.9\% to 67.1\% for size classification. On public logs from 4,061 additional drivers, the full-input model identifies powertrain correctly for 86 of 88 car models; size accuracy is 63.2\%, close to the internal vehicle-level result of 63.5\%. Physical-feature analysis links powertrain recognition to deceleration behaviour and size recognition to steering response. Errors near size-class boundaries and limited gains from longer observations highlight the importance of label resolution and training-fleet coverage for cross-model generalisation.
\end{abstract}

\section{Introduction}\label{sec:intro}

Vehicle trajectories support traffic analysis, simulation calibration and infrastructure planning. NGSIM~\cite{fhwa2016ngsim}, highD~\cite{krajewski2018highd} and I-24 MOTION~\cite{gloudemans2023motion} capture traffic interactions, often with dimensions or broad vehicle classes; powertrain information is less available, especially in anonymised probes. Inferring attributes could help describe fleet composition and calibrate vehicle behaviour. USDOT's \emph{Vehicle Classification Using Mobile Sensors} project distinguishes passenger cars and trucks from speed and acceleration~\cite{ban2013mobile}; FHWA also uses trajectories for microsimulation calibration~\cite{hale2021calibration}.

Wheelbase, steering ratio and mass distribution affect steering response; powertrain design affects acceleration and deceleration. Battery-electric vehicles (BEVs) can regenerate while internal-combustion-engine (ICE) vehicles coast. Driver behaviour, traffic and driving automation systems (DAS) also shape motion. We ask whether these signals reveal vehicle attributes on car models absent from training.

We classify three footprint-based sizes (wheelbase $\times$ mean track width) and BEV versus ICE using 60-s motion and CAN windows, including steering, pedal states and lead-vehicle interactions. Nearby footprints straddle class boundaries, automation changes motion, and models contribute uneven amounts of data. Testing on unseen models examines whether learned cues transfer.

Figure~\ref{fig:teaser} outlines VTT. Our contributions are:
\begin{enumerate}[label=(\arabic*)]
\item \textbf{Dataset and evaluation.} VTT provides labelled motion and CAN signals, fixed splits for unseen models and drivers, and window-, drive- and vehicle-level evaluation.
\item \textbf{Cross-model benchmark and method study.} We compare four sequence architectures, statistical and LLM-proposed features, and event learning. Training adaptations improve both tasks, with evaluation on model-disjoint folds and independent public logs.
\item \textbf{Physical interpretation and error analysis.} We connect recognition to deceleration and steering response, and assess automation, observation length, footprint boundaries and fleet coverage.
\end{enumerate}

\section{Related Work}\label{sec:related}

\textbf{Trajectory datasets.} NGSIM provides camera-derived trajectories at 0.1-s intervals~\cite{fhwa2016ngsim}. highD records naturalistic highway traffic from drones, including vehicle dimensions and classes~\cite{krajewski2018highd}. I-24 MOTION extends observation over a 4.2-mi freeway corridor using roadside cameras~\cite{gloudemans2023motion}. VTT complements them with onboard motion and control signals, platform-linked size and powertrain labels, and explicit tests on unseen car models. Its source logs come from the open-source driving-automation ecosystem used by comma2k19~\cite{schafer2018comma2k19} and ADAS-TO~\cite{wang2026adasto}. Public onboard datasets also support external checks: VED records vehicle energy and driving data~\cite{oh2019ved}, while OpenACC captures commercial adaptive-cruise-control behaviour~\cite{makridis2021openacc}.

\textbf{Vehicle attributes from motion.} GPS-based classifiers distinguish passenger cars and trucks using engineered features~\cite{sun2013gps} or convolutional networks~\cite{dabiri2020gps}. Naturalistic and instrumented studies report differences between electric and combustion vehicles in driver adaptation~\cite{helmbrecht2014ev}, adaptive-cruise-control operation~\cite{geffen2026ev} and regenerative-braking phases~\cite{liu2025paai}. We examine whether these differences support powertrain identification, alongside finer footprint-based size classes, on unseen car models.

\textbf{Identity cues and generalisation.} CAN signals can identify drivers from complete drives~\cite{enev2016fingerprinting} or individual turns~\cite{hallac2016turn}. If windows from the same driver or car model appear in training and testing, such cues can inflate type-recognition scores. This is a form of shortcut learning~\cite{geirhos2020shortcut}. VTT separates evaluation on unseen models from evaluation on unseen drivers of known models, and estimates uncertainty by resampling car models.

\section{Dataset and Evaluation}\label{sec:dataset}

\begin{figure*}[t]
\centering
\includegraphics[width=\textwidth]{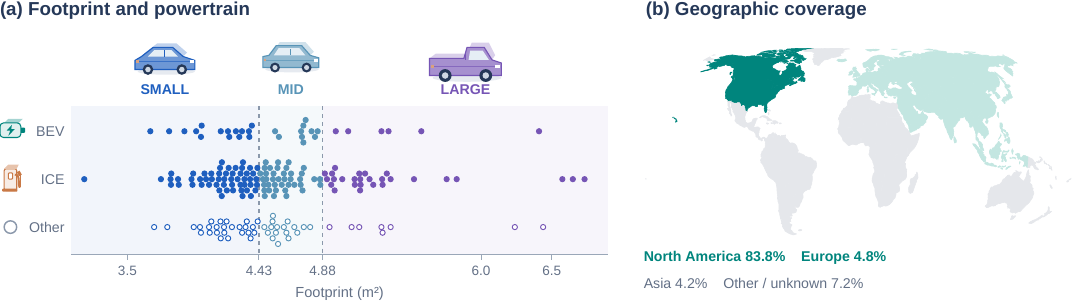}
\caption{VTT coverage. (a) Median footprints of 202 car-model--powertrain groups (167 models); dashed lines mark the size thresholds. Other includes hybrid, mixed and unresolved powertrains. Vehicle icons illustrate size classes. (b) Regional shares of driving hours; only coarse regions are retained.}
\label{fig:coverage}
\end{figure*}

\subsection{Source and Signals}
VTT uses public driving-automation logs from the ecosystem described in Section~\ref{sec:related}. Each drive spans one ignition cycle. Signals are aligned to the 10-Hz vehicle-state stream, with sample $k$ at nominal time $k/10$ s. Lower-rate radar, pose and controller signals use causal zero-order holding; stale values become missing, and gaps longer than 0.3 s are flagged. DAS periods are retained and marked.

Of 44 table columns, 20 are model inputs: speed and longitudinal acceleration; steering-wheel angle, rate and normalised torque; yaw rate, curvature and lateral acceleration; accelerator-pressed, brake-pressed, steering-pressed, standstill and DAS-engaged flags; and lead validity, distance, relative speed, speed, acceleration, time headway and time to collision. Lead acceleration is a causal slope of lead speed. Steering torque is divided by its drive-level 95th-percentile magnitude while moving and steering. Only pedal flags are used because accelerator and brake magnitudes are missing on 63.9\% and 49.4\% of drives. Lead estimates come from a camera model on 56.2\% of drives and from radar otherwise.

\subsection{Labels and Coverage}
Labels are assigned to the vehicle platform identified from CAN. A vehicle is a driver--platform pair. Size is footprint, defined as wheelbase $\times$ mean track width: SMALL below 4.43 m$^2$, MID from 4.43 to below 4.88 m$^2$, and LARGE at or above 4.88 m$^2$. The thresholds were chosen before classifier training. Each size $\times$ \{BEV, ICE\} cell required at least 100 h, three car models and five identity groups, linked by driver, vehicle identity or shared drive names. Among feasible threshold pairs, we maximised the smallest cell (309 h), placing cuts in gaps between platform footprints. BEV support across these classes comes from 15, 10 and 8 car models, compared with 55, 42 and 28 for ICE.

Powertrain labels are ICE, hybrid (HEV), plug-in hybrid (PHEV) and BEV; the main task is BEV versus ICE. Platforms with unresolved multiple powertrains or unknown powertrain have no powertrain label (11.1\% of drives). Table~\ref{tab:dataset} and Fig.~\ref{fig:coverage} summarise coverage.

\begin{table}[t]
\caption{Dataset coverage and class support.}
\label{tab:dataset}
\centering\footnotesize
\setlength{\tabcolsep}{4pt}
\renewcommand{\arraystretch}{1.05}
\begin{tabularx}{\columnwidth}{Lr}
\rowcolor{neutralband}\multicolumn{2}{l}{\bfseries\color{ink}Fleet and signal coverage} \\
Drives (one ignition cycle each) & 9,062 \\
Hours, nominal & 3,949.8 \\
Hours moving & 3,078.1 \\
Pseudonymous vehicles & 1,431 \\
Identity groups (unit of the splits) & 1,157 \\
Platforms / brands / car models & 360 / 38 / 174 \\
Samples with DAS engaged & 47.9\% \\
Samples with any automation activity & 62.6\% \\
\rowcolor{tablehead}\multicolumn{2}{l}{\bfseries\color{ink}Window availability} \\
Windows listed & 241,476 \\
Windows eligible & 201,390 \\
Windows human-only (strict) & 37,942 \\
Eligible windows, size & 201,169 \\
Eligible windows, BEV vs ICE & 144,534 \\
Eligible windows, four-class powertrain & 178,827 \\
\arrayrulecolor{ink}\bottomrule\arrayrulecolor{black}
\end{tabularx}
\vspace{3pt}
\begin{tabularx}{\columnwidth}{Lrrrr}
\tabhead\textbf{Class} & Windows & Hours & Vehicles & Platforms \\
\midrule
SMALL & 91,369 & 1,522.8 & 673 & 169 \\
MID & 56,059 & 934.3 & 392 & 104 \\
LARGE & 53,741 & 895.7 & 295 & 76 \\
\midrule
ICE & 79,566 & 1,326.1 & 598 & 200 \\
HEV & 23,708 & 395.1 & 197 & 57 \\
PHEV & 10,585 & 176.4 & 68 & 21 \\
BEV & 64,968 & 1,082.8 & 366 & 71 \\
Powertrain null & 22,563 & 376.1 & 156 & 34 \\
\arrayrulecolor{ink}\bottomrule\arrayrulecolor{black}
\end{tabularx}
\vspace{3pt}
\begin{tabularx}{\columnwidth}{Lrr}
\tabhead\textbf{Unseen-model test split} & Size & BEV vs ICE \\
\midrule
Eligible windows & 26,144 & 18,295 \\
Hours & 435.7 & 304.9 \\
Drives & 1,141 & 836 \\
Vehicles & 329 & 244 \\
Car models & 41 & 39 \\
\arrayrulecolor{ink}\bottomrule\arrayrulecolor{black}
\end{tabularx}
\vspace{3pt}
\tabnote{Vehicles are driver--platform pairs. Class counts refer to eligible 60-s windows; test counts cover the corresponding windows, drives, vehicles and car models.}
\end{table}

\subsection{Unseen-Model and Unseen-Driver Splits}
Both splits use identity groups to keep correlated drives together. The main split reserves unseen car models covering about 15\% of each size--powertrain cell's hours for testing. Groups that also drove a test model are excluded from the other splits. Validation groups are separate from training groups, but 53 of the 56 validation car models also occur in training. Model selection therefore uses the internal model-disjoint folds described in Section~\ref{sec:methods}.

The second split assigns groups within each cell to training, validation and testing in a 70/15/15 ratio by hours. Car models are shared, so it measures recognition for unseen drivers of known models. Research-fleet drivers remain in training.

\subsection{Prediction Units and Metrics}
An eligible 60-s window has 600 consecutive samples, no gap and at least 20\% moving samples. We evaluate all eligible windows, a published class-balanced subset, and a human-only subset with no automation activity during the window or the preceding 10 s. Drive predictions average window probability vectors before taking the most probable class; human-only drive predictions use only human-only windows. Vehicle predictions average all eligible windows across that vehicle's test drives. Units without a selected window are excluded.

We report macro-F1 over the classes present. Confidence intervals use 1,000 car-model cluster-bootstrap resamples (seed 0), keeping the observations of each sampled car model together.

\subsection{Metadata Checks and Data Access}
Group-disjoint checks with simple classifiers assess identity shortcuts~\cite{geirhos2020shortcut}. Nuisance metadata predict four-class powertrain at 0.32 macro-F1 and size at 0.40 (chance 0.25 and 0.33), so these columns are excluded. Statistics of the input signals yield brand 0.02 (chance 0.03 across 38 brands), platform 0.00, powertrain 0.31 and size 0.34, showing little identity or class information in this check.

Vehicles, drives and groups receive independent random pseudonyms. The release excludes positions, absolute times, dates, images, video and driver-monitoring signals; region is continent-level and time is relative to drive start. Access supports vehicle-dynamics and driving-behaviour research, with re-identification and linkage to other datasets prohibited. Data, indexes, split lists, schema and checksums use the \href{https://creativecommons.org/licenses/by/4.0/}{Creative Commons Attribution license}; build and evaluation code use the MIT License. Repository links are withheld for review.

\section{Methods}\label{sec:methods}
\subsection{Window Models and Baselines}
Each input is a $600\times20$ array, standardised with training-set channel statistics. We compare four sequence architectures~\cite{wang2017tsc,ismailfawaz2019review}: a one-dimensional ResNet~\cite{wang2017tsc}, two bidirectional GRU layers~\cite{cho2014gru}, a four-layer Transformer~\cite{vaswani2017attention}, and four bidirectional Mamba layers using state-space duality~\cite{gu2023mamba,dao2024mamba2}. The latter three share a two-layer convolutional stem with stride four (150 tokens, width 128); the Mamba implementation uses pure PyTorch. All use the same pooling-head design. Parameter counts are 2.31, 0.73, 0.70 and 1.07 million, respectively.

Baseline training uses class-balanced sampling with weights $(N/(Kn_k))^{0.5}$, capped at 10, where $N$ is the number of windows, $K$ the number of classes and $n_k$ the class count. We use AdamW~\cite{loshchilov2019adamw}, 3\% warm-up, cosine decay, batches of 256, label smoothing of 0.05 and early stopping on validation window macro-F1. Feature baselines use 180 statistics (nine per channel) and the physical features described below, separately or together, with class-balanced logistic regression and histogram gradient boosting~\cite{pedregosa2011sklearn,ke2017lightgbm}.

\subsection{Model Selection}
Training changes are selected on three internal folds per task drawn from the training and validation pool, with no car-model overlap within each fold. The size folds each reserve 19 of 57 candidate models; the powertrain folds reserve 16, 15 and 20 models. Training excludes all drives in identity groups associated with a fold's evaluation models. One pooled fleet group spans 19 car models and excludes only drives of evaluation models.

Baseline early stopping selects epochs 16--22 with the shared-model validation split, but epochs 1--8 (median about four) on model-disjoint folds. Adapted models use the median best epoch across folds: 11 and 16 for size (GRU and ResNet), two and seven for powertrain, and ten for the event model.

Selection criteria were fixed before the method comparison. Baseline fold means had to fall within five points of their test scores. Adapted size training targeted a fold score of 70; powertrain required 90 and a one-point improvement, or 92 on human-only drives. The event model required a two-point size gain over the best window model or a powertrain score of 91. Ensembles required a one-point gain over the best member; aggregation changes required one point on both tasks. Stored results were checked for reproducibility, and each selected model was evaluated once on the final test split.

\subsection{Training Adaptations}
For size, we successively add: car-model-balanced sampling within classes (sampling shares proportional to the square root of hours); a log-footprint regression head; ordinal classification~\cite{cao2020coral}; auxiliary log-curb-mass and four-class powertrain heads; and data augmentation. Augmentation jointly flips the signs of six lateral channels, masks one 3--15-s span while clearing lead validity, drops one channel group, and adds Gaussian noise of 0.02. From this augmented model we separately test same-class mixing of windows across car models, DAS weighting, conditional adversarial training~\cite{ganin2016dann,long2018cdan} with gradient reversal ($\lambda=0.1$), and per-car-model loss weighting~\cite{sagawa2020dro}.

Powertrain training follows the same sequence without ordinal classification or cross-model window mixing; its auxiliary heads also use HEV and PHEV windows. DAS weighting multiplies each window's loss by $(1-\beta)+\beta(1-s)$, where $s$ is its automation share and $\beta=0.5$. The selected size model uses per-car-model loss weighting; the selected powertrain model uses DAS weighting. Both predict the largest softmax probability.

\subsection{Event-Based Multiple-Instance Model}
The event model treats a drive as a collection of transient manoeuvres. We extract eight event types using Savitzky--Golay-smoothed speed~\cite{savitzky1964}: acceleration, deceleration with and without the brake pressed, stopping, starting, lead acceleration, lead deceleration and sustained low-speed turning. The extraction yields 735,571 events from 8,662 drives; 545 examined drives have no qualifying event. Each event is an 18-s onset-aligned segment with 22 channels (the inputs, smoothed acceleration and validity) and nine context values. Acceleration and deceleration are also represented as $a(v)$ on a 0--36 m/s speed grid with 0.5 m/s bins.

A shared width-128 encoder combines convolutions, residual blocks, event-type embeddings and a context multilayer perceptron. A second encoder processes $a(v)$; an adapter adds frozen 60-s window embeddings as a ninth instance type. Internal-fold evaluation uses that fold's own window checkpoints. Task-specific gated attention~\cite{ilse2018mil} pools instances for multiple-instance learning (MIL). Heads predict size, BEV--ICE, four-class powertrain, log footprint, log curb mass and two ordinal thresholds. The objective combines drive cross-entropies, instance cross-entropies (weight 0.3), footprint Huber loss (0.5), mass Huber loss (0.25), four-class cross-entropy (0.25) and ordinal binary cross-entropy (0.25). The model has about 0.99 million parameters and trains for ten epochs with AdamW, warm-up and cosine decay. We report its drive-level attention prediction and window-mean aggregation separately.

\subsection{LLM-Assisted Physical Features}
Large language models (LLMs) propose feature-extraction programs, following program-search approaches~\cite{romeraparedes2024funsearch}. Six proposing agents, each assigned a physical perspective, receive channel definitions and 2,727 training windows. They propose 74 features; six verification agents execute the code and assess class separation, removing 21 candidates. A further review removes 19, leaving 34 features. The minimum window coverage is reduced from 40\% to 10\% to retain features tied to less frequent driving conditions; gradient boosting can use their missingness. The feature set is fixed before evaluation. LLMs generate features, while the conventional classifiers make predictions.

\subsection{Aggregation}
The main results use the probability averaging defined in Section~\ref{sec:dataset}. Internal-fold comparisons also test DAS-, event- and steering-weighted arithmetic and log-odds means, robust and gated means, and a two-stage vehicle rule. A supplementary log-odds mean uses window weights $(1-s+0.1)^{0.5}(1+n_{\mathrm{acc}}+n_{\mathrm{dec}})$, where $n_{\mathrm{acc}}$ and $n_{\mathrm{dec}}$ count acceleration and deceleration events. This rule is fixed before final testing and reported alongside the selected models.

\section{Experiments}\label{sec:experiments}

Experiments use one RTX 5080 GPU (16 GB); gradient-boosting models use scikit-learn~\cite{pedregosa2011sklearn}. Macro-F1 is reported as a percentage, with drive-level performance on unseen car models as the main result. Table captions specify seeds; multi-seed results are mean $\pm$ standard deviation. Confidence intervals follow Section~\ref{sec:dataset}, averaging interval endpoints across seeds when applicable.

\subsection{Recognition on Unseen Models and Drivers}
The four baseline sequence models have similar scores (Table~\ref{tab:main}). Averaging windows within a drive adds about 7--12 points; further aggregation to vehicles offers no consistent gain. Human-only windows are generally harder for size and easier for powertrain. Human-only drive results cover fewer drives than the full set.

\begin{table*}[t]
\caption{Recognition on unseen car models: macro-F1 (\%).}
\label{tab:main}
\centering\footnotesize
\setlength{\tabcolsep}{2pt}
\renewcommand{\arraystretch}{1.13}
\begin{tabularx}{\textwidth}{>{\raggedright\arraybackslash}p{110pt}*{10}{C}}
\tabhead
 & \multicolumn{5}{c}{\bfseries\color{sizeblue}Size (SMALL / MID / LARGE)} & \multicolumn{5}{c}{\bfseries\color{powerteal}BEV vs ICE} \\
\cmidrule(lr){2-6} \cmidrule(lr){7-11}
 & \multicolumn{2}{c}{window} & \multicolumn{2}{c}{drive} & vehicle & \multicolumn{2}{c}{window} & \multicolumn{2}{c}{drive} & vehicle \\
\cmidrule(lr){2-3} \cmidrule(lr){4-5} \cmidrule(lr){6-6} \cmidrule(lr){7-8} \cmidrule(lr){9-10} \cmidrule(lr){11-11}
\tabhead\textbf{Model} & All & Human & All & Human & All & All & Human & All & Human & All \\
\midrule
Statistics + GBM & 49.7 & 46.3 & 54.7 & 50.5 & 51.5 & 69.5 & 73.5 & 73.2 & 74.3 & 76.0 \\
\rowcolor{rowtint}LLM features + GBM & 40.5 & 47.3 & 63.2 & 58.9 & 58.6 & 69.6 & 84.2 & 89.4 & 89.1 & 90.1 \\
LLM + statistics + GBM & 54.1 & 49.8 & 64.9 & 64.2 & 61.8 & 73.2 & 80.2 & 80.9 & 82.6 & 83.4 \\
\midrule
\rowcolor{rowtint}1-D ResNet & \res{53.6}{1.0} & 48.0 & \res{61.1}{1.3} & 60.3 & 56.4 & \res{78.4}{0.4} & 83.2 & \res{91.1}{0.8} & 89.5 & 90.3 \\
bi-GRU & \res{54.7}{0.3} & 50.4 & \res{62.9}{0.8} & 63.3 & 57.2 & \res{79.2}{0.8} & 85.2 & \res{89.0}{0.1} & 90.8 & 88.6 \\
\rowcolor{rowtint}Transformer & \res{52.5}{0.8} & 47.7 & \res{61.8}{0.9} & 61.3 & 55.3 & \res{77.0}{0.4} & 83.5 & \res{88.4}{0.6} & 90.1 & 89.0 \\
Mamba (bidirectional) & \res{53.1}{0.4} & 49.8 & \res{61.2}{0.3} & 61.1 & 55.2 & \res{77.3}{0.7} & 84.0 & \res{89.4}{0.3} & 90.9 & 90.6 \\
\midrule
\rowcolor{sizeband}bi-GRU, size training & \res{53.2}{0.3} & 47.8 & \best{\res{67.1}{0.0}} & 61.7 & \res{60.0}{0.6} & -- & -- & -- & -- & -- \\
\rowcolor{sizeband}ResNet, size training & \res{53.2}{0.8} & 49.5 & \res{66.8}{1.7} & 62.9 & \res{62.6}{1.4} & -- & -- & -- & -- & -- \\
\rowcolor{powerband}bi-GRU, powertrain training & -- & -- & -- & -- & -- & \res{77.0}{0.3} & 83.7 & \res{92.6}{0.1} & 89.1 & \res{93.4}{0.4} \\
\rowcolor{powerband}ResNet, powertrain training & -- & -- & -- & -- & -- & \res{79.9}{0.7} & 83.2 & \best{\res{93.0}{0.8}} & 88.7 & \res{92.8}{0.4} \\
\rowcolor{eventband}Event MIL, window mean & \res{54.2}{0.3} & 50.0 & \res{64.6}{0.5} & 62.5 & \res{62.3}{0.9} & \res{78.8}{0.1} & 82.2 & \res{90.2}{0.2} & 89.9 & \res{90.4}{0.5} \\
\rowcolor{eventband}Event MIL, bag prediction & -- & -- & 64.3 & 63.2 & 62.4 & -- & -- & \best{93.0} & 91.5 & 92.8 \\
\arrayrulecolor{ink}\bottomrule\arrayrulecolor{black}
\end{tabularx}
\tabnote{Size / BEV--ICE support: 41 / 39 car models, 1,141 / 836 drives (732 / 564 with human-only windows), and 329 / 244 vehicles. All: full evaluation set; Human: human-only subset. Sequence models use three seeds (standard deviations where available); gradient boosting (GBM) uses one. Chance is about 33 for size and 50 for powertrain. Coloured rows show task-specific training and event MIL; bold marks the best full-drive score for each task. Confidence intervals are reported in the text.}
\end{table*}

Adapted training improves drive-level macro-F1 by 4--6 points for size and 2--4 points for powertrain, without consistent window-level gains. For bi-GRU, size rises from 62.9 [48.9, 72.4] to 67.1 [54.4, 74.2], and powertrain from 89.0 [81.8, 93.7] to 92.6 [87.1, 96.2]. Gains vary little across seeds, although the car-model confidence intervals overlap. Event MIL reaches 64.3 for size and 93.0 for powertrain with attention aggregation; its powertrain advantage on internal folds does not carry to the test models.

Scores are higher for unseen drivers when car models are shared across splits (Table~\ref{tab:drivers}). LLM features plus statistics and the adapted ResNet both reach 80.6 for drive-level size. Adapted GRU size training lowers window-level performance but raises drive-level performance from 78.1 to 79.5 [70.8, 84.4], illustrating the difference between within-model recognition and cross-model training objectives.

\begin{table}[t]
\caption{Recognition on unseen drivers of known models: macro-F1 (\%).}
\label{tab:drivers}
\centering\footnotesize
\setlength{\tabcolsep}{2.3pt}
\renewcommand{\arraystretch}{1.12}
\begin{tabularx}{\columnwidth}{>{\raggedright\arraybackslash}p{83pt}*{4}{C}}
\tabhead
 & \multicolumn{2}{c}{\bfseries\color{sizeblue}Size} & \multicolumn{2}{c}{\bfseries\color{powerteal}BEV vs ICE} \\
\cmidrule(lr){2-3} \cmidrule(lr){4-5}
\tabhead\textbf{Model} & Window & Drive & Window & Drive \\
\midrule
\makecell[l]{LLM + statistics\\+ GBM} & 72.4 & 80.6 & 84.4 & 94.1 \\
\rowcolor{rowtint}1-D ResNet & \res{68.4}{3.0} & \res{77.4}{2.2} & \res{82.4}{0.7} & \res{96.2}{0.5} \\
bi-GRU & \res{70.2}{0.9} & \res{78.1}{1.4} & \res{87.4}{0.4} & \res{96.5}{0.1} \\
\rowcolor{powerband}ResNet, adapted & \res{64.8}{0.3} & \res{80.6}{0.1} & \res{83.1}{0.2} & \res{97.2}{0.3} \\
\rowcolor{powerband}bi-GRU, adapted & \res{62.9}{1.0} & \res{79.5}{0.7} & \res{78.8}{0.4} & \res{95.1}{0.3} \\
\rowcolor{eventband}Event MIL & 68.4 & 78.8 & 86.5 & 96.1 \\
\arrayrulecolor{ink}\bottomrule\arrayrulecolor{black}
\end{tabularx}
\tabnote{Size: 1,218 drives, 219 vehicles, 81 car models; BEV--ICE: 922 drives, 143 vehicles, 63 models. Full evaluation set. Mean $\pm$ standard deviation over three seeds; one seed for GBM and MIL. Adapted: task-specific training.}
\end{table}

\subsection{Training and Aggregation}
Baseline internal-fold means fall within five points of their unseen-model test scores (Table~\ref{tab:folds}). On the first size fold, training changes span only four points and none reaches the target of 70. Powertrain benefits most from augmentation; adding DAS weighting yields a three-fold mean of 93.1 for GRU. Event MIL matches the adapted size model and improves powertrain by 2.7 points on the internal folds. Its window-mean scores are lower (61.1 and 88.8). On the first fold, gated attention outperforms a learned instance mean: 67.8 versus 66.7 for size and 96.0 versus 94.8 for powertrain.

Shared-model validation scores differ substantially from unseen-model results: 60.7/62.8 for adapted GRU/ResNet size, 75.6/79.3 for powertrain, and 79.1/97.5 for event MIL size/powertrain. We therefore use the training durations selected on model-disjoint folds.

\begin{table}[t]
\caption{Training comparisons on model-disjoint internal folds: drive-level macro-F1 (\%).}
\label{tab:folds}
\centering\footnotesize
\setlength{\tabcolsep}{3pt}
\renewcommand{\arraystretch}{1.06}
\begin{tabularx}{\columnwidth}{Lcc}
\rowcolor{neutralband}\multicolumn{3}{l}{\bfseries\color{ink}(a) First-fold comparison, bi-GRU} \\
\tabhead\textbf{Training change} & \textcolor{sizeblue}{\textbf{Size}} & \textcolor{powerteal}{\textbf{BEV--ICE}} \\
Baseline & 65.3 & 89.4 \\
\rowcolor{rowtint}Model-balanced sampling & 62.8 & 88.0 \\
Footprint head & 66.8 & 87.4 \\
\rowcolor{rowtint}Ordinal head (size only) & 64.8 & -- \\
Mass and powertrain heads & 65.2 & 88.7 \\
\rowcolor{rowtint}Data augmentation & 66.2 & 92.8 \\
Cross-model window mixing & 64.8 & -- \\
\rowcolor{powerband}DAS weighting & 65.6 & 92.8 \\
Adversarial model head & 66.5 & -- \\
\rowcolor{sizeband}Per-model loss weighting & 66.9 & -- \\
\arrayrulecolor{ink}\bottomrule\arrayrulecolor{black}
\end{tabularx}
\vspace{4pt}
\begin{tabularx}{\columnwidth}{Lcccc}
\rowcolor{neutralband}\multicolumn{5}{l}{\bfseries\color{ink}(b) Three-fold evaluation} \\
\tabhead\textbf{Training} & Fold 1 & Fold 2 & Fold 3 & Mean \\
\rowcolor{sizeband}\multicolumn{5}{l}{\bfseries\color{sizeblue}Size} \\
GRU, baseline & 65.3 & 69.5 & 60.4 & 65.1 \\
\rowcolor{rowtint}GRU, footprint head & 66.8 & -- & -- & 65.5 \\
GRU, augmentation & 66.2 & -- & -- & 66.5 \\
\rowcolor{rowtint}GRU, model weighting & 66.9 & 70.5 & 64.8 & 67.4 \\
ResNet, baseline & -- & -- & -- & 66.6 \\
\rowcolor{rowtint}ResNet, model weighting & 67.0 & 70.4 & 61.2 & 66.2 \\
Event MIL, attention & 67.8 & 65.6 & 67.5 & 67.0 \\
\rowcolor{powerband}\multicolumn{5}{l}{\bfseries\color{powerteal}BEV versus ICE} \\
GRU, baseline & 89.4 & 90.8 & 95.4 & 91.9 \\
\rowcolor{rowtint}GRU, augmentation & 92.8 & -- & -- & 92.8 \\
GRU, DAS weighting & 92.8 & 91.1 & 95.4 & 93.1 \\
\rowcolor{rowtint}ResNet, baseline & -- & -- & -- & 93.1 \\
ResNet, DAS weighting & 95.7 & 93.9 & 96.1 & 95.2 \\
\rowcolor{eventband}Event MIL, attention & 96.0 & 98.2 & 93.2 & 95.8 \\
\arrayrulecolor{ink}\bottomrule\arrayrulecolor{black}
\end{tabularx}
\tabnote{Full evaluation set; one seed per entry. Changes in (a) accumulate through augmentation; subsequent rows branch from that model. Powertrain omits the ordinal head and window mixing. MIL uses its attention-based drive prediction. Dashes denote unreported fold scores.}
\end{table}

On internal folds with baseline GRU and ResNet predictions pooled, the best aggregation variant raises size from 67.5 to 67.8. DAS-weighted averaging raises powertrain from 93.0 to 94.8. Neither meets the one-point criterion on both tasks, so the main results retain plain averaging. A GRU--ResNet ensemble gains less than one point over the best member: 68.8 versus 68.2 for size and 93.9 versus 93.1--93.5 for powertrain. Stacking yields 66.5 and 92.3, respectively. On the six adapted runs per task, the supplementary weighted log-odds rule improves every run: size rises from 66.9 to 68.7 (about 1.7 points), and powertrain from 92.8 to 93.5 (+0.7).

\subsection{Error Analysis and Physical Cues}
\begin{figure}[t]
\centering
\includegraphics[width=0.88\columnwidth]{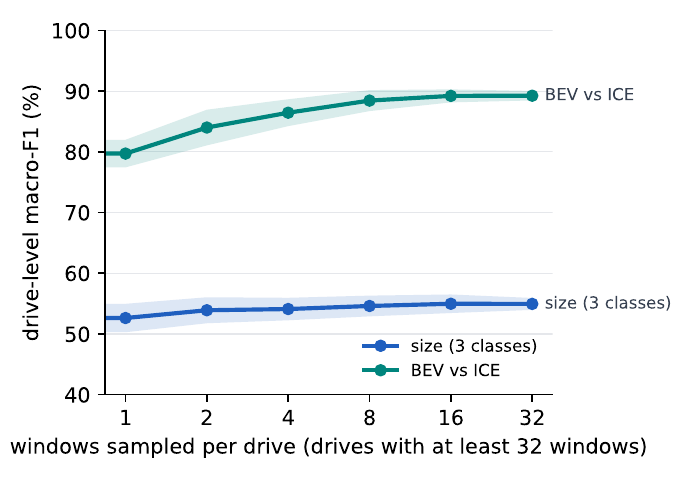}
\caption{Observation length on unseen models. For drives with at least 32 windows, predictions average $m$ sampled windows (50 draws; band: one standard deviation). Size macro-F1 rises from 52.6 at $m=1$ to 54.9 at $m=32$ (all windows: 56.0); powertrain rises from 79.7 to 89.3 (all: 89.1), levelling off by 8--16 windows.}
\label{fig:evidence}
\end{figure}

\begin{figure}[t]
\centering
\includegraphics[width=0.88\columnwidth]{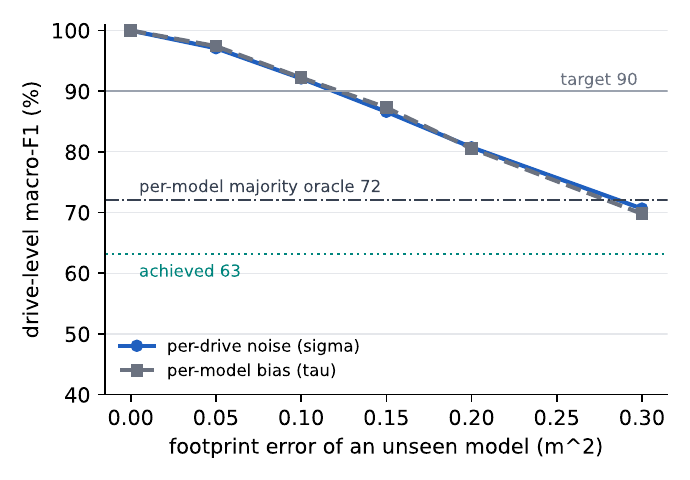}
\caption{Sensitivity to footprint error on unseen models (200 repeats). True footprint plus per-drive noise ($\sigma$, circles) or per-model bias ($\tau$, squares) is classified at 4.43 and 4.88 m$^2$. Horizontal lines show the baseline ensemble (63.2), its per-model majority prediction (72.0) and a target of 90.}
\label{fig:ceiling}
\end{figure}

Three diagnostic calculations examine size errors without further training. Assigning every drive of a car model its majority predicted class raises the ensemble of 12 baseline sequence runs from 63.2 to 72.0 for size and from 89.8 to 93.5 for powertrain. Thus, correcting inconsistent predictions within each model leaves substantial size errors. Adding Gaussian noise to the true footprint gives size macro-F1 of 92.1 at $\sigma=0.10$ m$^2$ and 86.6 at $\sigma=0.15$ m$^2$ (Fig.~\ref{fig:ceiling}). In this simulation, a score near 90 corresponds to a footprint error scale of about 0.12 m$^2$, or 2.5\% of a 4.6 m$^2$ footprint. Using true curb mass with two thresholds fitted on training drives gives 63.3 on 1,139 test drives with known mass.

Errors concentrate near class boundaries. Of 41 unseen models, 21 are within 0.15 m$^2$ of a differently labelled training model. In one baseline GRU run, window accuracy is 53.0\% within 0.1 m$^2$ of a threshold and 89.9\% beyond 0.8 m$^2$. Pickups and body-on-frame SUVs are often recognised as LARGE (GMC Yukon, Ford Ranger and Ford F-150 Lightning: 95.5--100\% of drives), while several large unibody models are assigned smaller classes (Genesis G80/G90, Volkswagen Atlas, Tesla Model S and Kia EV9: 0--5.6\% correct). MID models just above 4.43 m$^2$ are often classified as SMALL: GWM Haval H6, Audi Q5, Lexus RX and Nissan Altima have 0\% correct drives; Kia Stinger has 25\%.

Longer observation provides little additional size information for the baseline models (Fig.~\ref{fig:evidence}). Across drive-length groups from 1--2 to more than 64 windows, accuracy ranges from 63.8--73.8\% for size and 85.7--93.2\% for powertrain, without increasing with acceleration, deceleration or lead-acceleration event counts.

\begin{figure*}[t]
\centering
\includegraphics[width=0.9\textwidth]{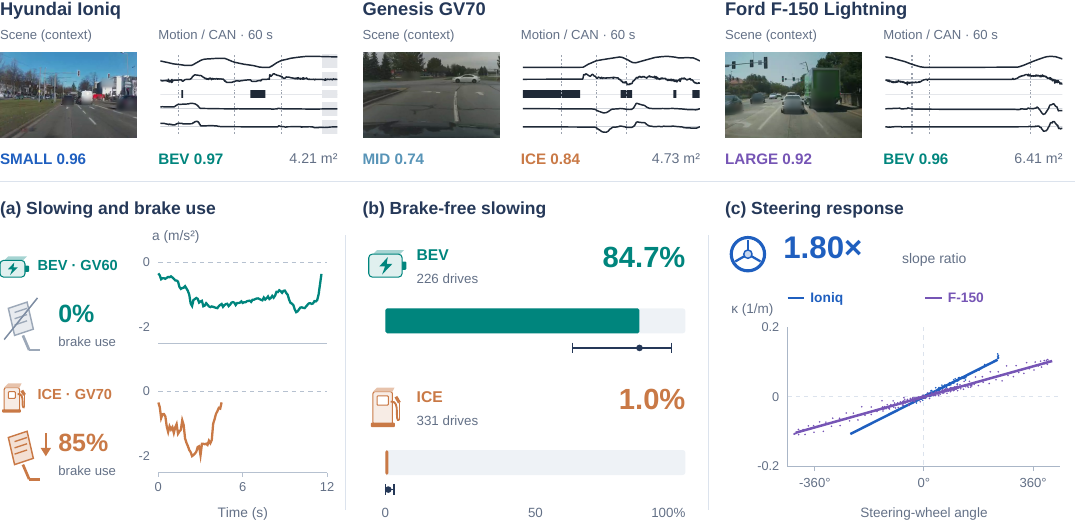}
\caption{Examples and physical cues. Top: three ego vehicles, 60-s signal windows and predicted-class probabilities. Camera frames provide context only; signal lanes show speed, acceleration, brake state, steering angle and yaw rate. Bottom: (a) acceleration in two slowing episodes, aligned at onset; percentages show brake use within each episode; (b) median per-drive brake-free slowing shares (bars) and interquartile ranges (whiskers), for $a<-1$ m/s$^2$, $v>3$ m/s and no DAS; (c) curvature versus steering angle at 2--12 m/s without DAS, with origin-constrained fits.}
\label{fig:examples}
\end{figure*}
The 34 LLM-proposed features perform better after drive-level aggregation, reaching powertrain scores close to the baseline sequence models (Table~\ref{tab:llm}). Combining them with statistics improves size but reduces drive-level powertrain performance. Validation-set permutation importance associates powertrain with lift-off deceleration, deceleration with the accelerator pressed or brake released, lead-to-ego deceleration delay, and the 10th percentile of brake-free deceleration. Size relies on curvature per steering angle at 6--18 m/s and low speed, Ackermann gain at 6--12 m/s, and deceleration delay.

Figure~\ref{fig:examples} illustrates these cues. For drives with at least 50 qualifying slowing samples and brake use recorded at least once, median brake-free slowing shares are 84.7\% for BEVs and 1.0\% for ICE vehicles. The Ioniq-to-F-150 steering-response slope ratio is 1.80, compared with a wheelbase ratio of 1.37; steering ratio as well as geometry contributes to the difference.

\begin{table}[t]
\caption{Physical features with gradient boosting: macro-F1 (\%).}
\label{tab:llm}
\centering\footnotesize
\setlength{\tabcolsep}{2pt}
\renewcommand{\arraystretch}{1.12}
\begin{tabularx}{\columnwidth}{>{\raggedright\arraybackslash}p{67pt}*{6}{C}}
\tabhead
 & \multicolumn{3}{c}{\bfseries\color{sizeblue}Size} & \multicolumn{3}{c}{\bfseries\color{powerteal}BEV vs ICE} \\
\cmidrule(lr){2-4} \cmidrule(lr){5-7}
\tabhead\textbf{Features} & All & Human & Drive & All & Human & Drive \\
\midrule
Statistics & 49.7 & 46.3 & 54.7 & 69.5 & 73.5 & 73.2 \\
\rowcolor{powerband}LLM physical & 40.5 & 47.3 & 63.2 & 69.6 & 84.2 & 89.4 \\
\rowcolor{rowtint}LLM + statistics & 54.1 & 49.8 & 64.9 & 73.2 & 80.2 & 80.9 \\
\arrayrulecolor{ink}\bottomrule\arrayrulecolor{black}
\end{tabularx}
\tabnote{Unseen car models, one seed. All and Human report window-level scores; Drive uses all windows. LLM physical: the 34 retained LLM-proposed features.}
\end{table}

\subsection{External Validation}
We apply the selected models to public logs without fine-tuning (Table~\ref{tab:external}). For datasets with fewer signals, models with the same training procedure are trained on the corresponding VTT subset: speed-only (speed, acceleration, standstill) or lateral plus speed (ten channels, excluding lead, torque and pedal flags). The accompanying code includes the full external evaluation.

commaCarSegments~\cite{commacarsegments} contains 45,763 one-minute windows (762.7 h) from 4,061 drivers absent from VTT, covering 113 car models and 161 openpilot platforms. Available CAN signals are decoded with opendbc. Yaw rate, curvature and lateral acceleration are derived from steering using a bicycle model; the understeer gradient $k=0.00079$ s$^2$/m$^2$ is fitted to IMU yaw on 62 VTT training and validation platforms. The full-input model classifies powertrain correctly for 86/88 labelled car models (97.7\% [96.2, 100.0]) and 87.8\% of 2,512 labelled drivers, rising to 88.9\% after yaw correction. Among 29 models from VTT's unseen-model test set, the 27 with powertrain labels are all classified correctly; driver accuracy is 94.2\% on 463 drivers.

With corrected yaw, size accuracy is 63.2\% and macro-F1 is 60.3, close to internal vehicle-level values of 63.5\% and 59.7. LARGE recall is 51.9\% externally versus 50.6\% internally. Pacifica, Palisade, CX-9, Escalade, fourth-generation Sienna and pickup platforms are recognised at platform level. A decoder correction recovers nine Honda platforms: 226/230 drivers are correctly classified as SMALL and 227/229 as ICE, including all five drivers of the unseen HR-V.

commaSteeringControl~\cite{commasteeringcontrol} provides 36 platform archives (51,462 highway windows, 857.7 h, 25 car models) under lateral control, with one archive treated as a vehicle. The speed-only model identifies powertrain correctly for 35/36 archives and 9/10 unseen-model archives. Lateral-plus-speed size accuracy/macro-F1 are 63.9\%/62.3, compared with 62.6\%/58.0 internally. With speed-only inputs, 258/264 ICE vehicles in VED~\cite{oh2019ved} and 99.1\% of 429 OpenACC trips~\cite{makridis2021openacc} are classified as ICE; the latter involve one Fiat 500X and 20 drivers. MoreTorque results are also reported in Table~\ref{tab:external}~\cite{moretorque}.

\begin{table}[t]
\caption{External recognition without fine-tuning: accuracy (\%) and correctly classified car models.}
\label{tab:external}
\centering\footnotesize
\setlength{\tabcolsep}{2.5pt}
\renewcommand{\arraystretch}{1.12}
\begin{tabularx}{\columnwidth}{>{\raggedright\arraybackslash}p{88pt}C>{\centering\arraybackslash}p{66pt}>{\centering\arraybackslash}p{27pt}}
\tabhead\textbf{Dataset / cohort} & \textbf{Inputs} & \textbf{Correct units} & \textbf{Models} \\
\rowcolor{powerband}\multicolumn{4}{l}{\bfseries\color{powerteal}BEV versus ICE} \\
\makecell[l]{commaSteeringControl\\36 archives} & Speed & 35 / 36 (97.2) & 24 / 25 \\
\rowcolor{rowtint}\makecell[l]{commaCarSegments\\2,512 labelled drivers} & All & 87.8 [86.6, 89.1] & 86 / 88 \\
\makecell[l]{commaCarSegments\\labelled drivers} & Speed & 92.6 [91.5, 93.6] & 87 / 88 \\
\rowcolor{rowtint}\makecell[l]{commaCarSegments\\463 drivers\\unseen models} & All & 94.2 [92.0, 96.2] & 27 / 27 \\
\makecell[l]{commaCarSegments\\229 drivers\\9 Honda platforms} & All & 227 / 229 (99.1) & 4 / 4 \\
\rowcolor{rowtint}MoreTorque, 11 ICE cars & All & 11 / 11 & -- \\
VED, 264 ICE cars & Speed & 258 / 264 (97.7) & -- \\
\rowcolor{rowtint}\makecell[l]{OpenACC, Fiat 500X\\429 trips} & Speed & 99.1 of trips & -- \\
\rowcolor{sizeband}\multicolumn{4}{l}{\bfseries\color{sizeblue}Size (SMALL / MID / LARGE)} \\
\makecell[l]{commaSteeringControl\\36 archives} & \makecell{Lateral\\+ speed} & 63.9 [50.0, 80.6] & 16 / 25 \\
\rowcolor{rowtint}\makecell[l]{commaCarSegments\\all 4,061 drivers} & All & 63.2 & 72 / 113 \\
\makecell[l]{commaCarSegments\\230 drivers\\9 Honda platforms} & All & 226 / 230 (98.3) & 5 / 5 \\
\rowcolor{rowtint}MoreTorque, 8 SMALL cars & All & 8 / 8 & -- \\
\arrayrulecolor{ink}\bottomrule\arrayrulecolor{black}
\end{tabularx}
\tabnote{All: 20 channels; Speed: speed, acceleration and standstill; lateral + speed: ten channels. Counts refer to the drivers, vehicles, archives or trips indicated; intervals are 95\% bootstrap CIs over evaluation units. commaCarSegments drivers are absent from VTT. Its all-driver size row uses corrected yaw; other rows retain their reported decoding. Macro-F1 is given in the text.}
\end{table}

\section{Discussion}\label{sec:discussion}

Size recognition is sensitive to small geometric errors near class boundaries. Similar scores across architectures and observation lengths suggest prioritising broader fleet coverage, continuous footprint estimation and wider class margins. For unseen drivers of known models, adapted methods reach 79.5--80.6 drive-level macro-F1.

DAS alters the motion--attribute relationship. Automated windows score higher for size (56.4 versus 50.2, baseline GRU) but lower for powertrain (79.8 versus 84.0, adapted ResNet) in individual runs. Model-specific controllers may contribute. Human-only results and DAS weighting examine this influence; model-disjoint folds better represent deployment on new models than shared-model validation.

The fleet comes from one logging ecosystem and is regionally uneven (Fig.~\ref{fig:coverage}). Incorrect platform identification affects every associated drive; 11.1\% lack a resolved powertrain label. The main results use onboard pedal and steering signals. Roadside or lead-vehicle applications require models trained on the available channels and tested under the target sampling and noise conditions. Speed-only external results provide an initial check.

\section{Conclusion}\label{sec:conclusion}
VTT benchmarks vehicle-attribute recognition across unseen car models. Powertrain macro-F1 reaches 92.6--93.0, while adapted size training improves bi-GRU from 62.9 to 67.1. Deceleration and steering provide physical cues; footprint and fleet coverage remain priorities for size recognition.

Applications include class-specific traffic summaries, microsimulation calibration and electrified-fleet analysis, extending USDOT/FHWA work~\cite{ban2013mobile,hale2021calibration}. Future work will broaden coverage, refine size labels and evaluate roadside and onboard trajectory inputs.

\bibliographystyle{IEEEtran}
\bibliography{refs}

@IEEEtranBSTCTL{IEEEexample:BSTcontrol,
  CTLdash_repeated_names = "no",
  CTLuse_forced_etal = "yes",
  CTLmax_names_forced_etal = "6",
  CTLnames_show_etal = "1"
}

@article{gu2023mamba,
  title   = {Mamba: Linear-Time Sequence Modeling with Selective State Spaces},
  author  = {Gu, Albert and Dao, Tri},
  journal = {arXiv preprint arXiv:2312.00752},
  year    = {2023}
}

@inproceedings{dao2024mamba2,
  title     = {Transformers are {SSM}s: Generalized Models and Efficient Algorithms Through Structured State Space Duality},
  author    = {Dao, Tri and Gu, Albert},
  booktitle = {Proc. Int. Conf. Mach. Learn. (ICML)},
  year      = {2024}
}

@inproceedings{vaswani2017attention,
  title     = {Attention Is All You Need},
  author    = {Vaswani, Ashish and Shazeer, Noam and Parmar, Niki and Uszkoreit, Jakob and Jones, Llion and Gomez, Aidan N. and Kaiser, {\L}ukasz and Polosukhin, Illia},
  booktitle = {Adv. Neural Inf. Process. Syst.},
  year      = {2017}
}

@inproceedings{cho2014gru,
  title     = {Learning Phrase Representations using {RNN} Encoder--Decoder for Statistical Machine Translation},
  author    = {Cho, Kyunghyun and van Merri{\"e}nboer, Bart and Gulcehre, Caglar and Bahdanau, Dzmitry and Bougares, Fethi and Schwenk, Holger and Bengio, Yoshua},
  booktitle = {Proc. Conf. Empirical Methods Natural Lang. Process. (EMNLP)},
  pages     = {1724--1734},
  year      = {2014}
}

@inproceedings{wang2017tsc,
  title     = {Time Series Classification from Scratch with Deep Neural Networks: A Strong Baseline},
  author    = {Wang, Zhiguang and Yan, Weizhong and Oates, Tim},
  booktitle = {Proc. Int. Joint Conf. Neural Netw. (IJCNN)},
  pages     = {1578--1585},
  year      = {2017}
}

@inproceedings{ilse2018mil,
  title     = {Attention-based Deep Multiple Instance Learning},
  author    = {Ilse, Maximilian and Tomczak, Jakub M. and Welling, Max},
  booktitle = {Proc. Int. Conf. Mach. Learn. (ICML)},
  pages     = {2127--2136},
  year      = {2018}
}

@article{cao2020coral,
  title   = {Rank Consistent Ordinal Regression for Neural Networks with Application to Age Estimation},
  author  = {Cao, Wenzhi and Mirjalili, Vahid and Raschka, Sebastian},
  journal = {Pattern Recognit. Lett.},
  volume  = {140},
  pages   = {325--331},
  year    = {2020}
}

@inproceedings{sagawa2020dro,
  title     = {Distributionally Robust Neural Networks for Group Shifts: On the Importance of Regularization for Worst-Case Generalization},
  author    = {Sagawa, Shiori and Koh, Pang Wei and Hashimoto, Tatsunori B. and Liang, Percy},
  booktitle = {Proc. Int. Conf. Learn. Represent. (ICLR)},
  year      = {2020}
}

@article{ganin2016dann,
  title   = {Domain-Adversarial Training of Neural Networks},
  author  = {Ganin, Yaroslav and Ustinova, Evgeniya and Ajakan, Hana and Germain, Pascal and Larochelle, Hugo and Laviolette, Fran{\c{c}}ois and Marchand, Mario and Lempitsky, Victor},
  journal = {J. Mach. Learn. Res.},
  volume  = {17},
  number  = {59},
  pages   = {1--35},
  year    = {2016}
}

@article{geirhos2020shortcut,
  title   = {Shortcut Learning in Deep Neural Networks},
  author  = {Geirhos, Robert and Jacobsen, J{\"o}rn-Henrik and Michaelis, Claudio and Zemel, Richard and Brendel, Wieland and Bethge, Matthias and Wichmann, Felix A.},
  journal = {Nat. Mach. Intell.},
  volume  = {2},
  number  = {11},
  pages   = {665--673},
  year    = {2020}
}

@article{pedregosa2011sklearn,
  title   = {Scikit-learn: Machine Learning in {P}ython},
  author  = {Pedregosa, Fabian and Varoquaux, Ga{\"e}l and Gramfort, Alexandre and Michel, Vincent and Thirion, Bertrand and Grisel, Olivier and Blondel, Mathieu and Prettenhofer, Peter and Weiss, Ron and Dubourg, Vincent and Vanderplas, Jake and Passos, Alexandre and Cournapeau, David and Brucher, Matthieu and Perrot, Matthieu and Duchesnay, {\'E}douard},
  journal = {J. Mach. Learn. Res.},
  volume  = {12},
  pages   = {2825--2830},
  year    = {2011}
}

@article{schafer2018comma2k19,
  title   = {A Commute in Data: The comma2k19 Dataset},
  author  = {Schafer, Harald and Santana, Eder and Haden, Andrew and Biasini, Riccardo},
  journal = {arXiv preprint arXiv:1812.05752},
  year    = {2018}
}

@article{wang2026adasto,
  title   = {{AD}{AS}-{TO}: A Large-Scale Multimodal Naturalistic Dataset and Empirical Characterization of Human Takeovers during {AD}{AS} Engagement},
  author  = {Wang, Yuhang and Xu, Yiyao and Sun, Jingran and Zhou, Hao},
  journal = {arXiv preprint arXiv:2603.06986},
  year    = {2026}
}

@article{savitzky1964,
  title   = {Smoothing and Differentiation of Data by Simplified Least Squares Procedures},
  author  = {Savitzky, Abraham and Golay, Marcel J. E.},
  journal = {Anal. Chem.},
  volume  = {36},
  number  = {8},
  pages   = {1627--1639},
  year    = {1964}
}

@inproceedings{loshchilov2019adamw,
  title     = {Decoupled Weight Decay Regularization},
  author    = {Loshchilov, Ilya and Hutter, Frank},
  booktitle = {Proc. Int. Conf. Learn. Represent. (ICLR)},
  year      = {2019}
}

@article{romeraparedes2024funsearch,
  title   = {Mathematical Discoveries from Program Search with Large Language Models},
  author  = {Romera-Paredes, Bernardino and Barekatain, Mohammadamin and Novikov, Alexander and Balog, Matej and Kumar, M. Pawan and Dupont, Emilien and Ruiz, Francisco J. R. and Ellenberg, Jordan S. and Wang, Pengming and Fawzi, Omar and Kohli, Pushmeet and Fawzi, Alhussein},
  journal = {Nature},
  volume  = {625},
  pages   = {468--475},
  year    = {2024}
}

@article{ismailfawaz2019review,
  title   = {Deep Learning for Time Series Classification: A Review},
  author  = {Ismail Fawaz, Hassan and Forestier, Germain and Weber, Jonathan and Idoumghar, Lhassane and Muller, Pierre-Alain},
  journal = {Data Min. Knowl. Discov.},
  volume  = {33},
  number  = {4},
  pages   = {917--963},
  year    = {2019}
}

@inproceedings{ke2017lightgbm,
  title     = {{LightGBM}: A Highly Efficient Gradient Boosting Decision Tree},
  author    = {Ke, Guolin and Meng, Qi and Finley, Thomas and Wang, Taifeng and Chen, Wei and Ma, Weidong and Ye, Qiwei and Liu, Tie-Yan},
  booktitle = {Adv. Neural Inf. Process. Syst.},
  year      = {2017}
}

@inproceedings{long2018cdan,
  title     = {Conditional Adversarial Domain Adaptation},
  author    = {Long, Mingsheng and Cao, Zhangjie and Wang, Jianmin and Jordan, Michael I.},
  booktitle = {Adv. Neural Inf. Process. Syst.},
  year      = {2018}
}

@article{helmbrecht2014ev,
  title   = {How Electric Vehicles Affect Driving Behavioral Patterns},
  author  = {Helmbrecht, Magnus and Olaverri-Monreal, Cristina and Bengler, Klaus and Vilimek, Roman and Keinath, Andreas},
  journal = {IEEE Intell. Transp. Syst. Mag.},
  volume  = {6},
  number  = {3},
  pages   = {22--32},
  year    = {2014}
}

@article{geffen2026ev,
  title   = {Safety, Mobility, and Environmental Impacts of Driver-Assistance-Enabled Electric Vehicles: An Empirical Study},
  author  = {Geffen, Gabriel and Zhao, Jun and Shang, Mingfeng and Wang, Shian and Wu, Yao-Jan},
  journal = {arXiv preprint arXiv:2601.17256},
  year    = {2026}
}

@article{liu2025paai,
  title   = {A Phase-Aware {AI} Car-Following Model for Electric Vehicles with Adaptive Cruise Control: Development and Validation Using Real-World Data},
  author  = {Liu, Yuhui and Wang, Shian and Panicker, Ansel and Embry, Kate and Asanova, Ayana and Li, Tianyi},
  journal = {arXiv preprint arXiv:2510.21735},
  year    = {2025}
}

@article{sun2013gps,
  title   = {Vehicle Classification Using {GPS} Data},
  author  = {Sun, Zhanbo and Ban, Xuegang (Jeff)},
  journal = {Transp. Res. C, Emerg. Technol.},
  volume  = {37},
  pages   = {102--117},
  year    = {2013}
}

@article{dabiri2020gps,
  title   = {A Deep Convolutional Neural Network Based Approach for Vehicle Classification Using Large-Scale {GPS} Trajectory Data},
  author  = {Dabiri, Sina and Markovi{\'c}, Nikola and Heaslip, Kevin and Reddy, Chandan K.},
  journal = {Transp. Res. C, Emerg. Technol.},
  volume  = {116},
  pages   = {102644},
  year    = {2020}
}

@article{enev2016fingerprinting,
  title   = {Automobile Driver Fingerprinting},
  author  = {Enev, Miro and Takakuwa, Alex and Koscher, Karl and Kohno, Tadayoshi},
  journal = {Proc. Privacy Enhancing Technol.},
  volume  = {2016},
  number  = {1},
  pages   = {34--51},
  year    = {2016}
}

@inproceedings{hallac2016turn,
  title     = {Driver Identification Using Automobile Sensor Data from a Single Turn},
  author    = {Hallac, David and Sharang, Abhijit and Stahlmann, Rainer and Lamprecht, Andreas and Huber, Markus and Roehder, Martin and Sosi{\v{c}}, Rok and Leskovec, Jure},
  booktitle = {Proc. IEEE Int. Conf. Intell. Transp. Syst. (ITSC)},
  pages     = {953--958},
  year      = {2016}
}

@misc{commasteeringcontrol,
  title        = {comma{S}teering{C}ontrol},
  author       = {{comma.ai}},
  year         = {2023},
  howpublished = {\url{https://huggingface.co/datasets/commaai/commaSteeringControl}}
}

@misc{commacarsegments,
  title        = {comma{C}ar{S}egments},
  author       = {{comma.ai}},
  howpublished = {\url{https://huggingface.co/datasets/commaai/commaCarSegments}}
}

@misc{moretorque,
  title        = {{MoreTorque}/rlogs},
  author       = {{MoreTorque}},
  howpublished = {\url{https://huggingface.co/datasets/MoreTorque/rlogs}}
}

@article{oh2019ved,
  title   = {Vehicle Energy Dataset ({VED}), a Large-Scale Dataset for Vehicle Energy Consumption Research},
  author  = {Oh, Geunseob and LeBlanc, David J. and Peng, Huei},
  journal = {arXiv preprint arXiv:1905.02081},
  year    = {2019}
}

@article{makridis2021openacc,
  title   = {{OpenACC}. {An} Open Database of Car-Following Data to Study the Properties of Commercial {ACC} Systems},
  author  = {Makridis, Michail and Mattas, Konstantinos and Anesiadou, Aikaterini and Ciuffo, Biagio},
  journal = {Transp. Res. C, Emerg. Technol.},
  volume  = {125},
  pages   = {103047},
  year    = {2021}
}

@misc{fhwa2016ngsim,
  author = {{Federal Highway Administration}},
  title = {Next Generation Simulation ({NGSIM}) Vehicle Trajectories and Supporting Data},
  year = {2016},
  howpublished = {Dataset, U.S. Department of Transportation},
  doi = {10.21949/1504477}
}

@inproceedings{krajewski2018highd,
  author = {Krajewski, Robert and Bock, Julian and Kloeker, Laurent and Eckstein, Lutz},
  title = {The {highD} Dataset: A Drone Dataset of Naturalistic Vehicle Trajectories on German Highways for Validation of Highly Automated Driving Systems},
  booktitle = {Proc. IEEE Int. Conf. Intell. Transp. Syst. (ITSC)},
  pages = {2118--2125},
  year = {2018},
  doi = {10.1109/ITSC.2018.8569552}
}

@article{gloudemans2023motion,
  author = {Gloudemans, Derek and Wang, Yanbing and Ji, Junyi and Zachar, Gergely and Barbour, William and Hall, Eric and Cebelak, Meredith and Smith, Lee and Work, Daniel B.},
  title = {{I-24 MOTION}: An Instrument for Freeway Traffic Science},
  journal = {Transp. Res. C, Emerg. Technol.},
  volume = {155},
  pages = {104311},
  year = {2023},
  doi = {10.1016/j.trc.2023.104311}
}

@techreport{ban2013mobile,
  author = {Ban, Xuegang (Jeff) and Sun, Zhanbo},
  title = {Vehicle Classification Using Mobile Sensors},
  institution = {University Transportation Research Center},
  year = {2013},
  month = apr,
  url = {https://rosap.ntl.bts.gov/view/dot/25846}
}

@techreport{hale2021calibration,
  author = {Hale, David K. and Li, Xiaopeng and Ghiasi, Amir and Zhao, Dongfang and Khalighi, Farnoush and Aycin, Murat and James, Rachel M.},
  title = {Trajectory Investigation for Enhanced Calibration of Microsimulation Models},
  institution = {Federal Highway Administration},
  number = {FHWA-HRT-21-071},
  year = {2021},
  month = aug,
  url = {https://rosap.ntl.bts.gov/view/dot/57959}
}

\end{document}